\documentclass[a4paper,11pt]{article}         

\usepackage{graphicx}
\usepackage{mathptmx}      
\usepackage{amsmath}  
\usepackage{amsfonts}  
\usepackage{graphicx}
\usepackage{array} 
\usepackage{enumerate} 
\usepackage[ansinew]{inputenc} 
\usepackage[T1]{fontenc} 
\usepackage{url}
\usepackage[usenames]{color}
\usepackage{lscape}
\usepackage{placeins} 
\usepackage{rotating} 
\usepackage{fancyhdr}
\fancyheadoffset{30pt} 
\fancyfootoffset{30pt} 

\newcommand\unio{\mathop\cup\limits}
\newcommand\diss{\mbox{\rm diss}}
\newlength\savedwidth
\newcommand\whline{\noalign{\global\savedwidth\arrayrulewidth
                            \global\arrayrulewidth 1 pt}%
                   \hline
                   \noalign{\global\arrayrulewidth\savedwidth}}         
\begin{document}  

\title{Classifying the typefaces of the Gutenberg 42-line bible
}


\author{Aureli Alabert \\ 
           Department of Mathematics \\
           Universitat Aut\`onoma de Barcelona \\  
           08193 Bellaterra, Catalonia \\  
           \url{Aureli.Alabert@uab.cat}  
           \and
           Luz Ma. Rangel \\ 
           Department of Design and Image \\  
           Universitat de Barcelona \\  
           Pau Gargallo, 4 \\
           08028 Barcelona, Catalonia \\  
           \url{quadrati@luzrangel.com}
}

\date{\today}
\thispagestyle{empty}
\maketitle

\begin{abstract}  
We have measured the dissimilarities among several printed characters of a single page in the
Gutenberg 42-line bible and we prove statistically the existence of several different 
matrices from which the metal types where constructed. This is in contrast with the
prevailing theory, which states that only one matrix per character was used in the printing process
of Gutenberg's greatest work.

The main mathematical tool for this purpose is \emph{cluster analysis}, combined
with a statistical test for outliers. We carry out the research with two letters, \texttt{i} and \texttt{a}.
In the first case, an exact clustering method is employed; in the second, with more specimens    
to be classified, we resort to an approximate agglomerative clustering method. 

The results show that the letters form clusters according to their shape, with significant shape differences among clusters, 
and allow to conclude, with a very small probability of error, that indeed the metal types
used to print them were cast from several different matrices.

\end{abstract}

\section{Introduction}\label{Intro}

  It has been accepted for a long time that the typefaces of the 
  Gutenberg 42-line bible were produced by metal types coming from
  a unique \emph{matrix}. More precisely, Zedler \cite{Zedler}
  classified 299                                                                         
  different \emph{glyphs}, and this classification have been usually took as 
  definitive until today. 
  The 42-line bible (known for short as B42)  
  is commonly believed to be the  
  first printing work in which movable types were used, conferring
  Gutenberg the honour of the invention of this technology.

  However, when considering the personal background of Johannes Gutenberg 
  and of the other
  people involved in the project (notably Johann Fust and Peter Sch÷ffer), 
  together with 
  their historical circumstances and other technological reasons,
  one is compelled to question this single-matrix theory, and to 
  consider the possible existence of multiple matrices. 
  
  In this paper, we show statistical evidence of the existence 
  of more than one matrix to found the types. To this end, 
  we quantify numerically the dissimilarities between pairs of
  printed letters representing the same glyph, and
  we see that the letters cluster in a natural way in groups with
  shapes that are similar inside each group and significantly different
  from the shapes of the members in the other groups. 
  \emph{Statistical cluster analysis} is our main technique, 
  further
  complemented with \emph{outlier testing} to validate the clusters
  obtained. 
  
  The printed letter has some ``errors'' or ``deviations'' with respect to the ideal 
  shape derived from the metal type, due to inking or to paper inhomogeneity. 
  Moreover, metal types
  derived from the same matrix may present deviations with respect to the ideal 
  intended shape
  defined by that matrix (manufacturing deviations or wear).
  These errors can be considered as random, and independent for each printed letter. 
  We need to ``filter out'' these random deviations to observe if there is an actual 
  structural  
  difference among letters, attributable to the fact that they indeed come from
  different matrices.
  
  Obviously, letters printed with the same metal type do come from the same matrix 
  and will be very similar in shape. Therefore, we want to study a set of letters for which 
  we can ensure that all of them were printed 
  with different types. 
  If different letter shapes are then observed, we will
  know that these differences are due only to a diverse matricial origin.
  
  To guarantee that all types are different, we only need to take letters
  printed on the same page, since a whole page (two columns) was printed at once from a complete \emph{form} 
  made by the typographic composer. In contrast, opposite pages in the same sheet could not be printed concurrently
  by Gutenberg presses,
  so it would be wrong to use a set of letters coming from both pages. 
  
\begin{figure*}
\centering
{\includegraphics[scale=0.6, angle=270]{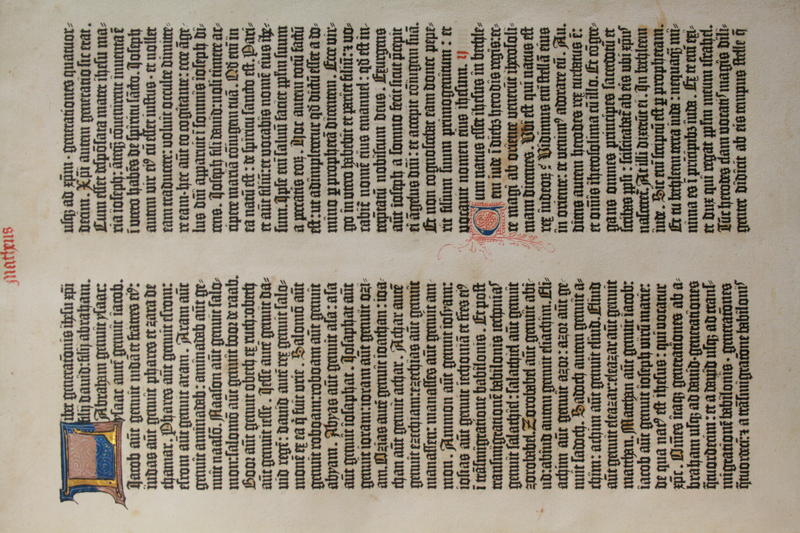}}  
\caption{\small The first page of the Gospel of Matthew of the New Testament kept at the Universidad de Sevilla, 
                 from were the letters that we compare were taken.}\label{Matthew1}
\end{figure*}

  We have chosen for convenience the first page of the Gospel of Matthew 
  (see Figure \ref{Matthew1})
  and we have considered in that page 10 letters \texttt{i} and 21 letters \texttt{a}. 
  The quantity of 10 objects is small enough to permit the computation of the optimal clustering 
  under any prefixed criterion, by simple exhaustive search.
  On the other hand, 21 objects allow to find out more classification patterns, although 
  we must resort to an approximate method of cluster analysis 
  because of the enormous time that the full enumeration would take.
  
  The paper is organised as follows: In Section \ref{metod} we introduce our main mathematical tools, 
  cluster analysis and outlier testing, recall some basic concepts of classical typography, and 
  briefly highlight the facts that confer its importance to the book under 
  consideration. Section 
  \ref{lab} is devoted to the detailed description of the procedure
  followed to measure the \emph{distances} between pairs of printed letters,
  which is the initial datum to apply the cluster analysis.
  Section \ref{clustersI} explains our exact method to find a good clustering and its application to 10 letters
 \texttt{i} in Matthew's first page; \emph{exact} means that an optimality criterion is established 
  and that the absolute optimum is found with respect to that criterion. The approximate cluster analysis
  is described in Section \ref{clanaprox}, and it is illustrated with letters \texttt{i}
  one more time. The results are compared to those obtained with the exact method; this comparison will permit us
  to ensure that the approximate method applied to the twenty-one letters \texttt{a}
  in the next Section
  \ref{clusterA} is meaningful. 
  The actual application of the hierarchical clustering is done in Subsection
  \ref{dendrosA}, whereas in \ref{validclust} we apply a statistical method of 
  detection of anomalous data (outliers) to validate the clusters obtained in \ref{dendrosA}.
  
  It is not possible to represent graphically each letter as a point in some space, in such 
  a way that the ordinary Euclidean distances among the points coincide exactly 
  with the measured distances,
  since the latter are not true distances in the mathematical sense (the triangle property does not hold true).
  However, it is possible to represent the relative positions of the letters approximately, in 2 and 3 dimensions,
  by means of the so-called  
  \emph{Multidimensional Scaling}. In Section \ref{MDS}, a representation is the plane is given for both letters.   
  The final Section \ref{conclusion} sketches the future plans in the line of research of this paper.
  
  The present study has a necessary conservative bias. We are asking the experimental data to give us 
  more than circumstantial evidence of the existence of several concurrent matrices, 
  since this is a conclusion that runs against the currently accepted theory.
\section{Preliminaries}\label{metod}

\subsection{Mathematical tools}
  \emph{Cluster analysis} comprises a variety of methods that intend to 
  obtain a reasonable grouping of a set of objects, based on the similarities
  among them. Each group is called a \emph{cluster}, and a particular 
  grouping, where each object is assigned to one cluster, is called a  
  \emph{clustering}.
  In modern statistics, cluster analysis is viewed as the main tool in the field 
  of \emph{unsupervised learning}.
   
  To perform a cluster analysis it is enough to have a measure representing a ``distance''
  or dissimilarity between each pair of objects.    
  Ideally, a cluster must possess  \emph{internal cohesion} (i.e. small dissimilarities 
  among the objects of the same cluster) and \emph{external isolation} (that means, large 
  dissimilarities between two objects of different clusters). 
  There are several reasonable ways to implement this idea.  
  The choice of one or another
  is a matter of mathematical modelling, and must be done to suit the
  particular situation at hand. In Subsection \ref{valclustering}
  we explain different good standard possibilities and justify our specific choice,
  which is in fact a variant of one of them, and seems to be new.

  Cluster analysis will be complemented, in the study of letters 
  \texttt{a}, with two additional tools. 
  On the one hand, we will use the data supplied by a small set 
  of letters printed with brand new types cast with the same matrix. 
  This set will play the role of control group to estimate which sort of variability 
  one can expect from letters with a common matricial origin
  (Subsection \ref{dendrosA}). On the other hand, 
  we will use statistical tests for the detection of extreme data (\emph{outliers}) 
  to confirm, with a probability of error controlled and small, that two 
  different clusters must be indeed considered distinct
  (Subsection \ref{validclust}).
  
  A classical exposition of cluster analysis can be found for instance in  
  Gordon \cite{Gordon}. A modern approach, imbedded in the so-called  
  \emph{unsupervised learning theory}, is Chapter
  14 of Friedman--Hastie--Tibshirani \cite{Friedman}. 
  The basic reference on outliers is the book by Barnett and Lewis \cite{Barnett}.  
  
\subsection{Typographical terms}  
  For the reader not well acquainted with the concepts of classical (i.e. non-digital)
  typography, we briefly explain some terms and procedures here. They are not strictly needed 
  to follow the rest of
  the article, but they help in understanding the motivation of our work.
  For a detailed and illustrated description, we refer the reader to Smeijers \cite{Smeijers}.
  
  A \emph{matrix} is a piece of brass or copper in which the letter is engraved, by
  means of a previously made steel tool called a \emph{punch}, which represented the
  character in reverse. One punch is needed for every character of a typeface and every size.  
  Some parts of the punch are actually made by means of a \emph{counterpunch}, but we do not
  need to go that far here.  

  The matrix already contains information on the height (\emph{point size}) 
  of the printed character, but the space that the metal type will really occupy 
  in the line (the \emph{character width}) is not yet determined. 
  The type itself is cast from the matrix and a \emph{mould} in an alloy of lead. 
  There is one degree of freedom in the position of the mould, 
  which determines the character width. 
  In the matrix, the character 
  has the right-reading direction, but it is engraved. In the type it   
  shows in the wrong-reading direction and in relief. 
  
  The types are then gathered in a \emph{form} to print a whole page. Several types of
  each character are therefore needed at the same time to compose a page. 
  
  The precise methods used by Gutenberg are still the subject of much research,
  but, according to \cite{Baines}, the method described here was well established by 1470.

\subsection{The 42-line bible}  
  
  The printing of the so called \emph{42-line bible} started in 1453 and was a 
  very serious enterprise at that time. Gutenberg had been working for years 
  perfecting the foundry mould technique with his previous metallurgical knowledge,
  and he needed the aid of the businessman 
  Johannes Fust to finance the project,  
  and the collaboration of the calligrapher Peter Sch÷ffer to develop the typographical 
  concept. The society was broken in the
  middle of the production of the approximately 200 copies of the book, and 
  Gutenberg was removed from the project and sued by Fust. The whole printing was completed 
  in August 1456. 
  
  There were other minor printing works before and after this one, and several 
  versions of the same manuscript are named, when possible, according to the 
  number of lines per page. The B42, however, possess several pages 
  printed at 40 lines, and it is not yet clear what was the reason. 
  It is anyway considered 
  the start of an era, because of the huge dimension of the work. Moreover, all copies
  show still today an extremely beautiful and bright ink, made from a recipe that  
  has been lost, and that was never used again in subsequent work. Chemical non-destructive
  analysis carried out in the 1980's (Cahill-al. \cite{Cahill}) have found out a high metallic 
  content, especially 
  of lead and copper, but the compounds used and their proportions are not known.
  
  All these facts have contributed to make this bible one of the most interesting 
  books of all times, from many scientific, social and historical points of view. 
  Currently, around 55 copies are known to have survived either totally or partially.
  An inventory can be found in the fundamental work of Paul Schwenke \cite{Schwenke}. For a general 
  history of the book, we recommend \cite{DeHamel}.

\section{How the dissimilarities were measured}\label{lab}
  Although we will use both terms \emph{distance} and \emph{dissimilarity},
  we recall that the dissimilarities here need not respect the triangular
  inequality, hence they are not distances in the usual mathematical sense.
  
  The measuring device was the Mitutoyo QVA-200 \cite{Mitutoyo}, with a resolution 
  of 0.0001 mm, and equipped with its proprietary software QVPack and FormPack. 
  The letters were measured from a back up microfilm of the New Testament
  of the 42-line bible located in the library of the Universidad de Sevilla
  \cite{microfilm}.
  
  To estimate the measurement error attributable to the device and the measuring process,
  we measured the dissimilarity of one scanned letter with itself, and the values were in the order 
  of $10^{-27}$ mm$^2$. Two different scans of the same letters gave rise to a value
  of at most $10^{-6}$. 
  As we will see, this value is at least one order of magnitude below the relevant values
  in measuring the dissimilarities between two different printed letters, and therefore 
  the measuring process can be considered stable and trustable in this aspect.
  This process is subsequently described.
  
  The measuring device scans each printed letter and determines the contour of its shape. 
  A contour is approximated by points and by line segments joining the points, forming a polygonal closed curve.
  Some letters possess empty inner spaces (\emph{counters}), and in this case the contour is constituted by 
  more than one connected curve. 
  
  The first step in the measuring process of two given contours $A$ and $B$ is   
  to place them in a coordinate plane, in a similar orientation and sharing a common 
  reference point, e.g. the barycenter. Then an initial distance between the two shapes is 
  determined and one of the shapes is moved in the 
  plane (with translations and rotations), in order to obtain a lower distance. 
  The distance is again computed and a new movement takes place. This iterative process
  stops when no further improvement seems possible. 
  
  At each iteration $n$, the distance is computed in the following way: From each point
  $p$ of contour $A$, its Euclidean distance $d(p,B)$ to contour $B$ is computed, that means, the distance
  to the closest segment of contour $B$. The squares of the distances are then added up:
\begin{equation*}
  D_n(A,B):=\sum_{p\in A} d(p,B)^2
\end{equation*}  
  If the process stops
  after $n_f$ iterations, the value $D_{n_f}$, divided by the number of points in $A$, 
  is taken as the dissimilarity $D(A,B)$ of 
  $A$ with respect to $B$. Since the role of both shapes in this process is not
  symmetric, the roles are interchanged and the corresponding quantity $D(B,A)$ is 
  computed. The final (symmetric) value of dissimilarity between $A$ and $B$ will be 
  the mean 
\begin{equation*}
  \diss(A,B):={\textstyle\frac{1}{2}}\big(D(A,B)+D(B,A)\big)
  \ .
\end{equation*}  
  The division by the number of points of the shape allows to compare different pairs of shapes
  in a common scale.
\section{Exact clusterings for letter \texttt{i}} \label{clustersI}    
\subsection{Dissimilarity table}

  The 10 letters \texttt{i} printed in the odd numbered lines in the first page
  of the Gospel of Matthew have been compared in pairs with the procedure
  described in Section \ref{lab}, and a symmetric table of dissimilarities has
  been obtained. This is the suitable initial datum  to proceed with the cluster
  analysis.
  
  The original dimensions of the letters in the 
  microfilm are in the orders of $10^{-1}$ mm (the 'x' height is around 0.285 mm), 
  and give rise to dissimilarities
  in the order of  10$^{-5}$ mm$^2$. For a more comfortable visualisation of the values,
  these have been multiplied by $10^4$, and the leading zeroes are omitted. The results are 
  depicted in Table \ref{diss_i}.
  
\begin{table*}
\centering
{\ttfamily
\begin{tabular}{r|rrrrrrrrrr}
        &i1     &i2     &i3    &i4     &i5     &i6     &i7     &i8     &i9    &i10
        \\
\hline
i1  &       &.2329 &.3518 &.2310 &.1912 &.3213 &.2179 &.2652 &.2590 &.3929
        \\
i2  &.2329 &       &.2097 &.0283 &.0701 &.0801 &.1139 &.0425 &.0835 &.0907
        \\
i3  &.3518 &.2097 &       &.1750 &.1638 &.3713 &.0670 &.1901 &.1232 &.3290
        \\
i4  &.2310 &.0283 &.1750 &       &.0895 &.0756 &.0902 &.0512 &.1026 &.0926
        \\
i5  &.1912 &.0701 &.1638 &.0895 &       &.1541 &.1219 &.1285 &.1028 &.2014
        \\
i6  &.3213 &.0801 &.3713 &.0756 &.1541 &       &.1467 &.0560 &.1324 &.0645
        \\
i7  &.2179 &.1139 &.0670 &.0902 &.1219 &.1467 &       &.0977 &.0900 &.2186
        \\
i8  &.2652 &.0425 &.1901 &.0512 &.1285 &.0560 &.0977 &       &.0718 &.0719
        \\
i9  &.2590 &.0835 &.1232 &.1026 &.1028 &.1324 &.0900 &.0718 &       &.1717
        \\
i10 &.3929 &.0907 &.3290 &.0926 &.2014 &.0645 &.2186 &.0719 &.1717 &      
\end{tabular}
}
\caption{\small Dissimilarity table for the \texttt{i}}  \label{diss_i}
\end{table*}

  In the next subsections, we will study the classification of the letters \texttt{i} starting 
  from this table.
  At the same time, we introduce the necessary cluster analysis theory. 
  The code to perform the statistical analysis have been written in 
  the language \texttt{R} \cite{RProject}. 
  
 \subsection{How to value each possible clustering} \label{valclustering}
 
  As mentioned in Section \ref{metod}, we intend to have \emph{cohesive} 
  and \emph{isolated} clusters. Suppose we have $N$ objects
  (in our case, the $N=10$ letters \texttt{i}), and assume that we want to divide these 
  objects
  into a given number $K$ of clusters. (Actually, 
  we do not want to fix the number of clusters from the start,
  but let us assume that we do.
  Later we come back to the discussion on the number of clusters.)
  
  In order to find the best clustering with $K$ clusters, we need to assign to each
  such clustering a certain cost, and then try to find the clustering with the minimum cost. 
  There are several reasonable possibilities to translate
  the desired properties of cohesion and isolation into a cost function.
  
  Common criteria proceed by evaluating the cost
  of each specific cluster and then combine the individual costs of the clusters.
  Both things can be done in several ways. For the first one, 
  the following values are typically used:
  \begin{enumerate}[1.]
  \item
    The maximum of the dissimilarities among objects in the cluster. 
  \item
    The sum of those dissimilarities. 
  \item \label{starsum} 
    For each object, the sum of all dissimilarities between the object and the 
    remaining ones are measured. Among all these quantities, the smallest one is
    chosen.
  \end{enumerate}
  Valuations based in the minimum or the sum of dissimilarities between objects in 
  the cluster and objects not in the cluster are also used.
  
  We will use a variant of criterion \ref{starsum}, whose justification will be seen later:
  \begin{enumerate}[4.]
  \item  
    For each object, the mean of all dissimilarities between the object and all other objects
    in the cluster is measured. Among these quantities, the smallest one is taken.
  \end{enumerate}
  
  To combine the costs of each individual cluster, there are also two usual criteria:
  
  \begin{enumerate}[(a)]
  \item
  Adding up the costs of all clusters.
  \item
  Taking the maximum of the costs.
  \end{enumerate}
                                                                                               
  Methods 3 and 4 have the advantage that they distinguish a particular object in each cluster,  
  namely the one for which the sum or mean of dissimilarities with the other objects is smallest
  (the same object in both methods, obviously).
  This allows to see the other objects as located 
  ``around the privileged one''. In our case, it allows to take one letter as a ``model''
   and to think of the other ones as ``variants'' of the model, representing, in theory, 
  a printed letter coming possibly from the same matrix as the model but with different metal types and  
  printing errors.
  
  Between possibilities (a) and (b), we prefer the second, because with the first 
  we may find clusters with a high cost coexisting with clusters of small cost.
  Taking the maximum as the value to minimize tends to make the clusters uniform in this sense,
  and in our opinion this is more reasonable and conservative in the situation we are studying.
  
  Once we decided to use criterion (b) for the combination of the costs of all clusters,
  the new proposed alternative 4 is better than the standard idea 3, 
  as the next example shows: Suppose that for the letters 
  \texttt{i} we want to compare clustering 
$$  
  \{1\}\ \ \{2,3,4,5,6,7,8,9,10\}
  \ ,
$$
  with clustering 
$$  
\{1,10\}\ \ \{2,3,4,5,6,7,8,9\}  
\ .
$$
  With the combination 3-b, the first one has a cost 
  of 0.7050 and the second one a cost of 0.6124, therefore making the second option preferable.
  But this does not seem reasonable, because letters
  1 and 10 are the farthest apart in the whole set. What happens here is that in the 
  first clustering we have a cluster of zero cost (because there is only one object in it, so that
  the internal cohesion is absolute)
  and another one with cost 0.7050; in the second clustering, the first cluster new cost
  is 0.3929, the distance between letters 1 and 10, and the second lowers to 0.6124. 
  The maximum has therefore dropped from 0.7050 to 0.6124, and the second clustering turns
  out to be better. 
  
  This inconvenience tends to appear with criteria 3-b whenever 
  we have clusters
  with a very unbalanced quantity of elements. With the alternative 4-b, the number of elements in 
  the clusters is not relevant. In the same example, now the first clustering has a cost of 
  0.0881, much better than the second, which is 0.3929.

  In summary, we take combination 4-b as the best suited for this study, 
  although we have not been able to find in the literature an example of 
  application of this particular combination of criteria. 
  Our choice gives rise to:
  
  \begin{itemize}
  \item
  The possibility of graphically representing the clusters 
  as stars, with a model letter in the center, and the other letters
  around, as variants of the model.
  \item
  No coexistence of very cohesive clusters together with clusters with little cohesion.  
  \item
  No coexistence, within a cluster, of letters distant from the center of the star, 
  together with letters close to it.
  \end{itemize}
  
\subsection{Finding the optimal clustering}\label{clustopt}  
  When the number of objects is small, and still assuming that we have fixed the 
  number of clusters, the optimal clustering 
  can be found by exhaustive enumeration of all possible clusterings. 
   
  We have used this direct method to classify the 
  10 letters \texttt{i} in $2$ and $3$ clusters.
  From the 511 possible partitions into two clusters, the best one is
$$  
  \{1\}\ \ \{2,3,4,5,6,7,8,9,10\}
  \ ,
$$
 whereas among the 9330 clusterings with three clusters one finds
 $$\{1\}\ \ \{3,7\}\ \ \{2,4,5,6,8,9,10\} \ .$$ 

\begin{figure}
\centering
{\includegraphics[scale=0.7]{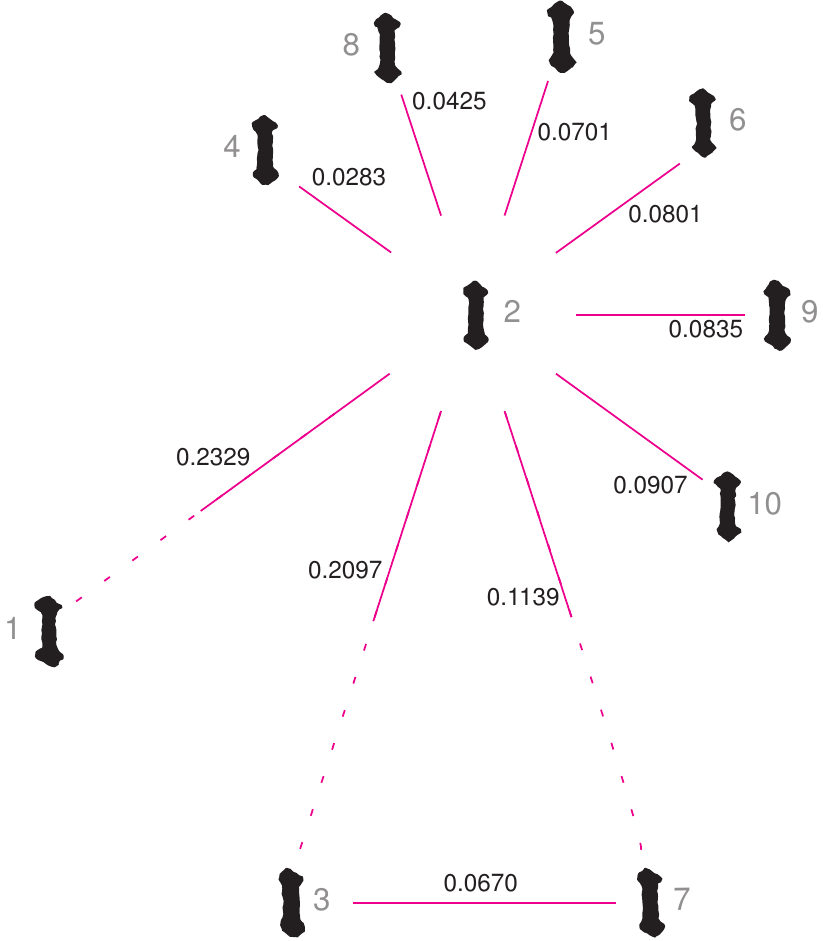}}  
\caption{\small A star-shaped representation of the best clustering with three clusters. The objects are  
reproductions of the actual letters scanned.}\label{stars_i_K3}
\end{figure}

  As mentioned above, it is possible to graphically represent these clusterings by means 
  of ``stars''. 
  The best clustering with three clusters is so displayed in Figure 
  \ref{stars_i_K3}. Letter \texttt{i1} is isolated, letters \texttt{i3} and    
  \texttt{i7} form a second cluster, in which it is irrelevant which letter is
  taken as the center, and finally the remaining letters form a cluster around
  letter \texttt{i2}. The distances from \texttt{i2} to its companion letters
  are shown on the solid lines, and the distances with the other three letters
  are also shown, on half-dashed lines.
  (The lengths of the lines in the picture are not proportional to the dissimilarity.)

  Notice that the internal distances are smaller than the distances among model letters
  of the different clusters. This is not a necessary consequence of the  
  criterion chosen, but and additional good property enjoyed by this particular clustering. 
  
  We could have tried to find the best clustering with four or more clusters, but 
  there are very few objects, and its validity would be questionable. 
  At the end of Subsection \ref{validclust} we give another reason against the possibility 
  of more than three clusters.
  
  For the 21 letters \texttt{a}, the direct and exact method was not computationally feasible, and we resorted 
  to and approximate method, as we will see in Sections \ref{clanaprox} and \ref{clusterA}.
\subsection{The number of clusters}
  
  The issue of the number of clusters present, or of the existence of more than one, does not have a universal
  solution. Our point of view is based in the following considerations: 
  
\begin{itemize}
\item
  First of all, we want to be in the conservative side; that means, we prefer to err by ascribing two letters
  to the same matrix when they actually come from different matrices, than to err by saying that they come 
  from different matrices when they in fact have a common origin.
\item  
  Secondly, our goal is not to determine exactly how many distinct matrices
  were used (even in only one page). We only want to decide if there is enough evidence that there were more than one, 
  or more than two, etc. 
\item  
  In the third place, although it is necessary to take some decisions a priori that may influence the result (choosing 
  a criterion for what is an optimal partition, or choosing the approximate method to employ), then,  
  consistently with the conservative philosophy, the clusters are reexamined with an additional criterion to ensure 
  its isolation (see Subsection \ref{validclust}).
\end{itemize}  

\section{Approximate clusterings} \label{clanaprox} 
  
  To compute the best partition into clusters, there is no way essentially more efficient 
  than the exhaustive enumeration
  that we have applied to letters \texttt{i}. Therefore,  
  if the enumeration is impossible because of the input size,
  one has to settle for some approximate method,
  able to provide a reasonably good solution.

  The approximate method that we use here corresponds to the heuristic procedure called 
  \emph{agglomerative hierarchical clustering} (see, e.g. \cite{Friedman}, \cite{Gordon}). 
  With this method, one constructs in fact a hierarchy of clusterings, so that
  one has to decide afterwards which clustering  in the hierarchy to keep.
  
  It starts by declaring each object as a cluster by itself
  and then bigger clusters are constructed from smaller ones in sequence. At each step, 
  two clusters are
  merged together to form a bigger one, namely, those which are 
  the closest in some sense. 
  
  This idea raises the need to define a notion of \emph{dissimilarity among clusters}, which involves again
  a certain arbitrariness in the choice. There are three reasonable and commonly used possibilities:
  
\begin{enumerate}
\item
  The dissimilarity between two clusters is the dissimilarity between their most similar objects.
  (This option is known in the literature as \emph{single linkage}.)
\item  
  The dissimilarity between two clusters is the dissimilarity between their most distant objects
  (known as \emph{complete linkage}).
  \item  
  The dissimilarity between two clusters is the arithmetic mean of the dissimilarities between the objects 
  in both clusters (called \emph{average linkage}).
\end{enumerate}  
  
  We illustrate the hierarchical agglomerative clustering arising from the three possibilities
  above with the letters 
  \texttt{i}, already studied with an exact method, and we discuss the correlation of each variant
  with respect to the exact results. We are able in this way to justify the choice of the next section 
  for the study 
  of the 21 letters  \texttt{a}.
    
\subsection{The \emph{single linkage}}  
\begin{figure*}
\centering
{\includegraphics[scale=0.6]{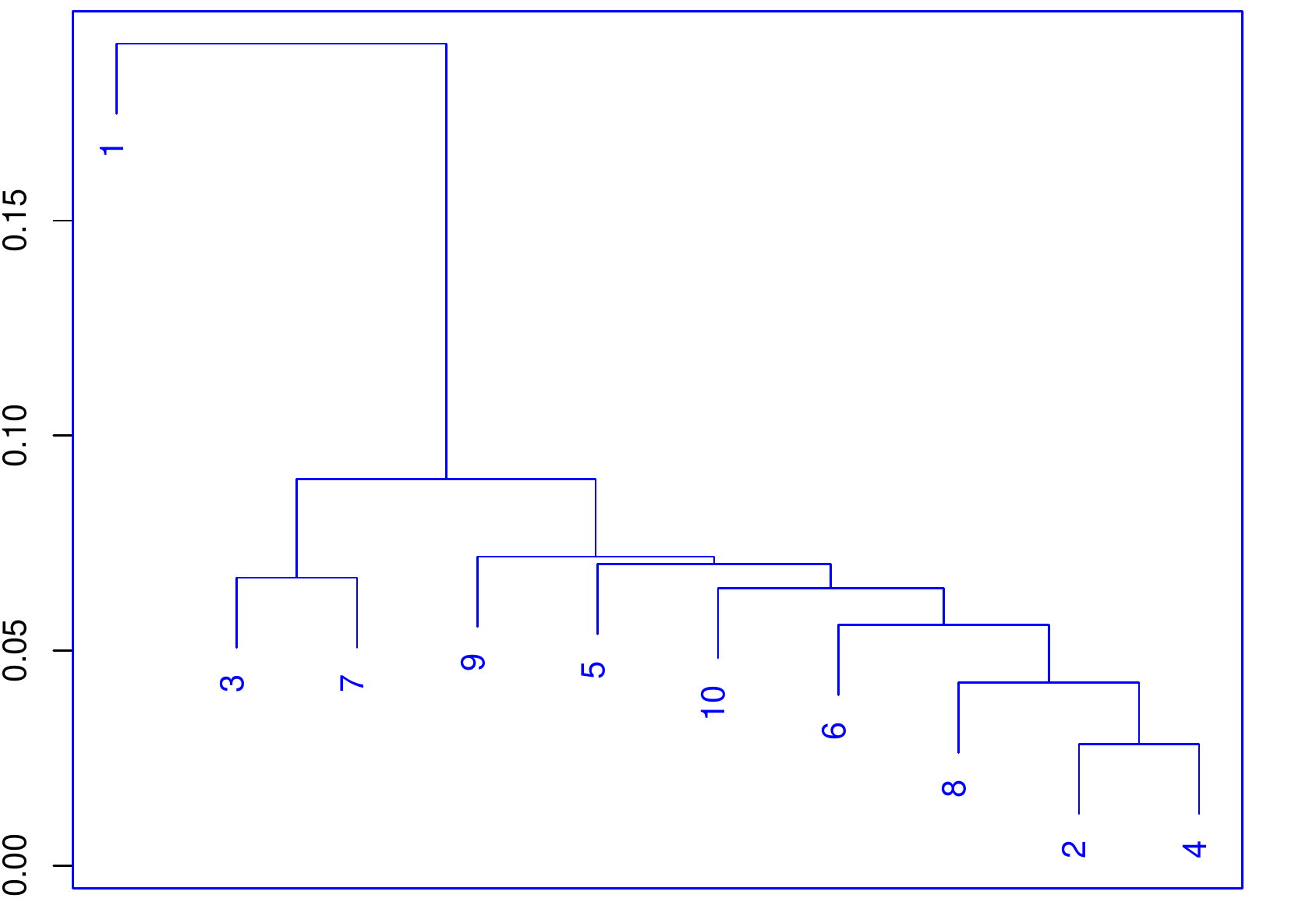}}  
\caption{\small Dendrogram for letter \texttt{i} with the \emph{single linkage} method}\label{hclust_i_single}
\end{figure*}

  With any of the three possibilities one can draw a graphical representation of the agglomerative
  process, called \emph{dendrogram}. For instance,
  the single linkage for the 10 letters \texttt{i} produces the dendrogram
  of Figure \ref{hclust_i_single}. The height of the horizontal lines, read in the vertical axis, 
  is the distance between 
  the two clusters that hang from the line, and that combine into one at that point.
  
  Now, looking at the dendrogram, we should decide how many and which clusters   
  seem to exist. When applying the method to the letters \texttt{a}, 
  we will use additional information to help us in this purpose. Here, we only want 
  to observe to which extent the approximate methods give similar results
  to the exact method used in the previous section.

  Specifically, we see in the dendrogram that letter 1 must be clearly set apart 
  from the others, and we also observe a clear separation between group $\{3,7\}$
  and the remaining letters, although not so clear as with letter 1.
  This reflects fairly well in a graphical way the results
  that we obtained when we imposed 2 or 3 clusterings. 
  Moreover, looking for a splitting in more than three clusters
  will not be supported by the dendrogram. 
  
  The single linkage method is very conservative, in the sense that it can easily consider  
  in the same cluster two very distant objects, as long as they are connected by a chain 
  of other objects, each one similar enough to the next.

\subsection{The \emph{complete linkage}}  
  
  The complete linkage method takes the opposite heuristic and this makes it little 
  conservative: It tends to keep separated pairs of objects that perhaps are not so different.
  The clusters produced tend to be cohesive but not quite isolated.  
  
\begin{figure*}
\centering  
{\includegraphics[scale=0.6]{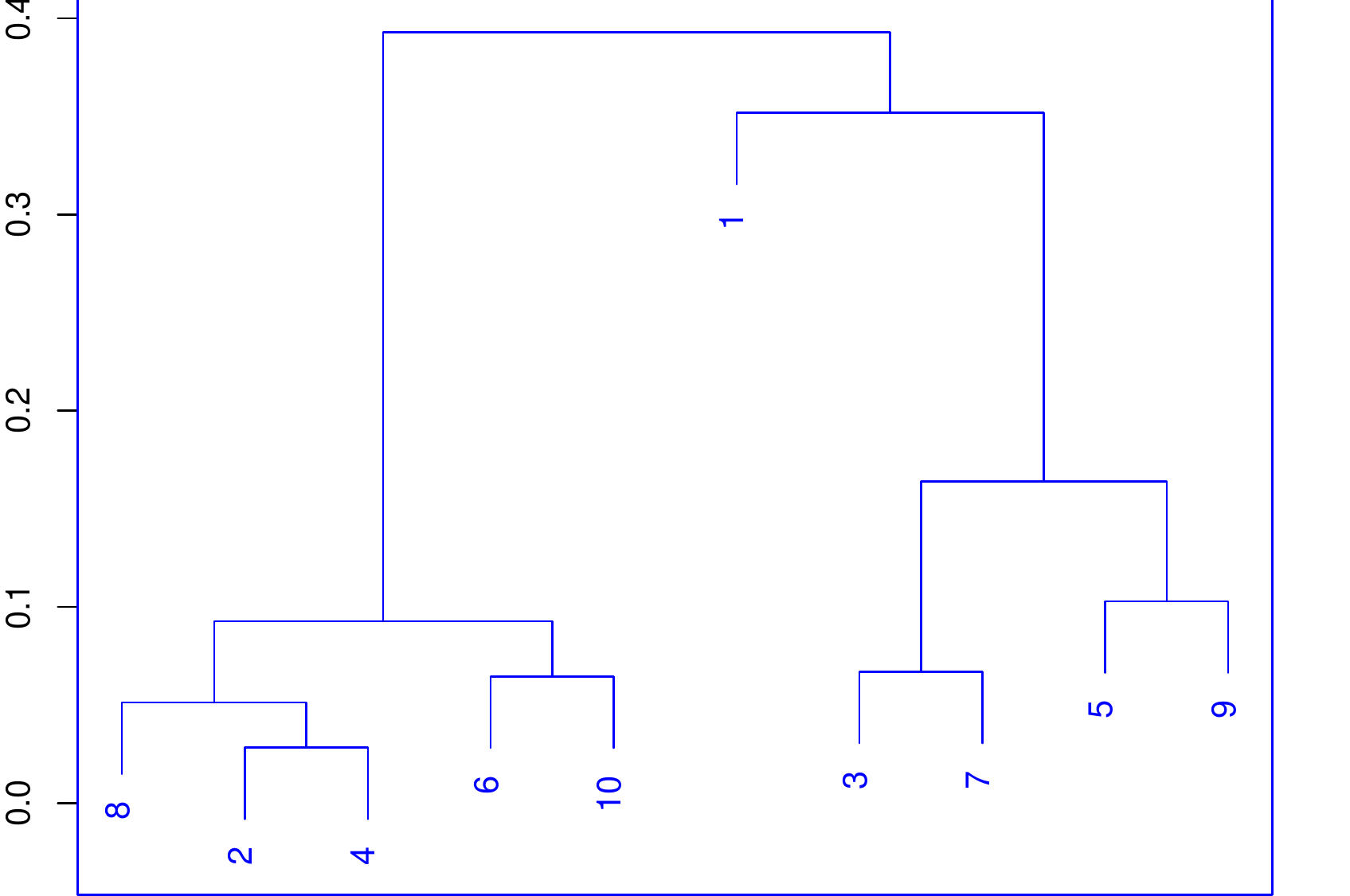}}  
\caption{\small Dendrogram for letter \texttt{i} with the \emph{complete linkage} method}\label{hclust_i_complete}
\end{figure*}

  With the complete linkage agglomerative clustering we obtain the dendrogram in Figure \ref{hclust_i_complete}. 
  Up to four clusters could be concluded from the picture. Moreover, even sticking
  to three clusters, the result would not coincide with that of the single linkage  
  (letters 5 and 9 switch clusters). 
  This dendrogram thus suggest a result more distant to the exact one than that obtained with the single linkage method.
  
\subsection{The \emph{average linkage}}  

\begin{figure*}
\centering  
{\includegraphics[scale=0.6]{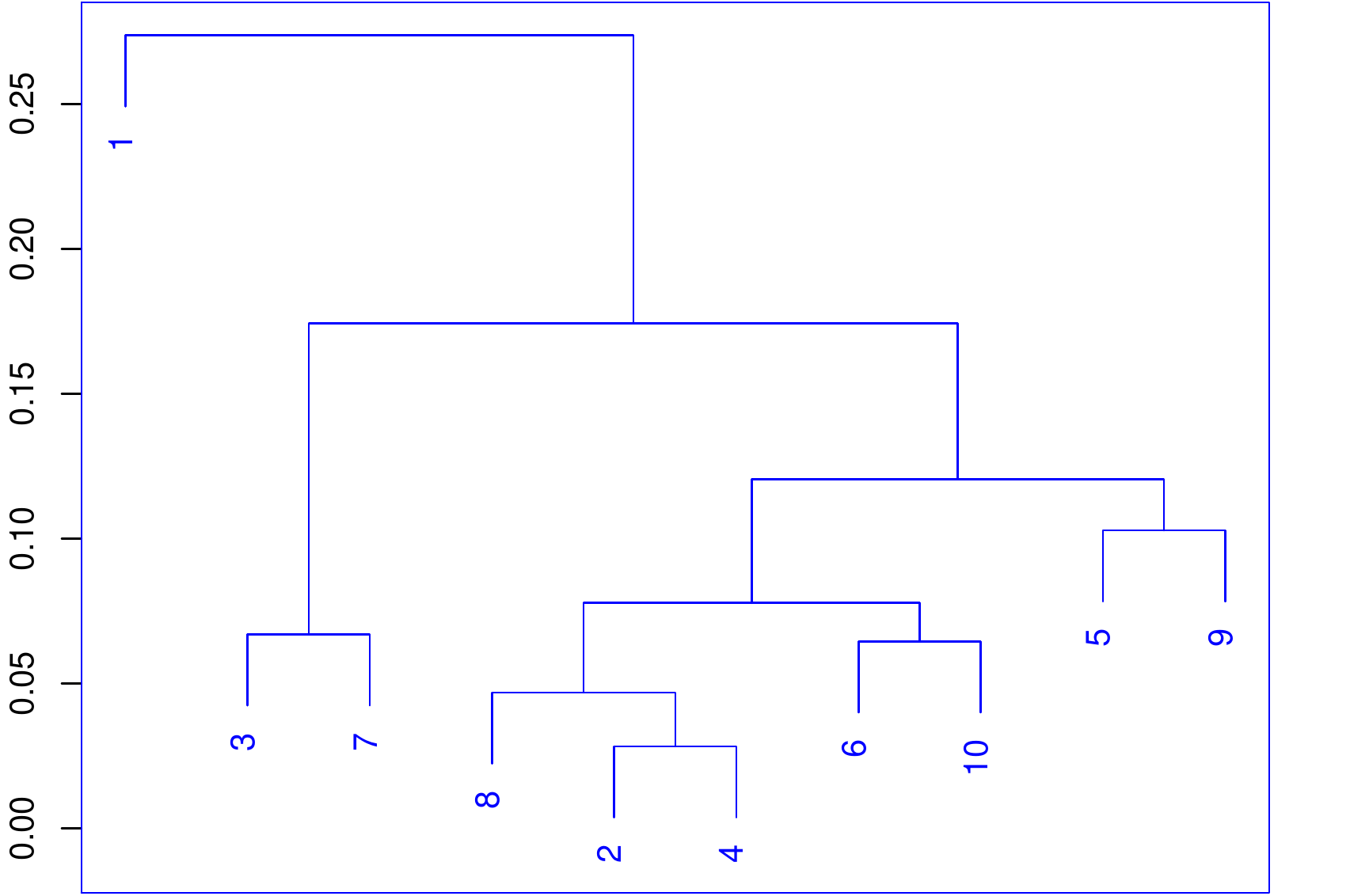}}  
\caption{\small Dendrogram for letter \texttt{i} with the \emph{average linkage} method}\label{hclust_i_average}
\end{figure*}

  The average linkage is half-way between the previous methods. 
  It gives rise to dendrogram of Figure \ref{hclust_i_average}. The distance 
  between two clusters $C_1$ and $C_2$ is computed as 
\begin{equation*}
  \frac{1}{|C_1|\cdot|C_2|}\sum_{o_1\in C_1}\sum_{o_2\in C_2} \diss(o_1, o_2)  
\end{equation*}  
  where $|C|$ means the number of elements of cluster $C$, and $\diss(o_1,o_2)$
  is the given dissimilarity between objects $o_1$ and $o_2$.  
  
  The following list indicates the order in which the clusters are formed, and the distance between
merging clusters. The central column indicates which clusters are merged. A number \texttt{n} refers to a 
single object (the n-th letter \texttt{i}), whereas a letter means the multiobject cluster formed in the 
line labelled with that letter.
\par
\medskip
\begin{center}
{\ttfamily
        \setlength{\extrarowheight}{2pt}
      \begin{tabular}{c|c|c|}
       \cline{2-3} 
         & {\rmfamily merge}  & {\rmfamily distance} \\
      \whline
       \multicolumn{1}{|c|}{a} & 2 -- 4  &  0.0283 \\
       \multicolumn{1}{|c|}{b} & 8 -- a  &  0.0469 \\
       \multicolumn{1}{|c|}{c} &~6 -- 10 &  0.0645 \\
       \multicolumn{1}{|c|}{d} & 3 -- 7  &  0.0670\\
       \multicolumn{1}{|c|}{e} & b -- c  &  0.0778\\
       \multicolumn{1}{|c|}{f} & 5 -- 9  &  0.1028\\
       \multicolumn{1}{|c|}{g} & e -- f  &  0.1205\\
       \multicolumn{1}{|c|}{h} & d -- g  &  0.1744 \\
       \multicolumn{1}{|c|}{i} & 1 -- h  &  0.2737 \\
      \hline                                                          
       \end{tabular}
  }
\end{center}

\par\medskip
We observe that the dendrogram here looks more correlated with the exact result
of Section \ref{clustersI} than the complete or even the single linkage dendrogram.
The clearest cut point produce the same 
partitions in two and three clusters and, moreover, the separation of the group
$\{3,7\}$
is more apparent here. Therefore we believe that this method is the one with results
more similar to the exact method corresponding to criterion  4-b that we have applied. 
We will adopt it for the classification of the twenty-one letters 
\texttt{a} in the next section.
  
\section{Cluster analysis for letter \texttt{a}}\label{clusterA}

  In the previous Section \ref{clanaprox}, we have compared several approximate  
  clustering methods with the exact cluster analysis performed with our optimality
  criterion. We concluded that the average linkage method was the one with the best
  results. We thus apply this method to classify  the twenty-one \texttt{a} taken 
  from the first page of the Gospel of Matthew.
  The corresponding dissimilarity measures can be seen in Table \ref{diss.a.completas}. 
\begin{table*}
\centering  
{\ttfamily
       \setlength{\extrarowheight}{2pt}
       \renewcommand{\tabcolsep}{3pt}
\scalebox{.60}[.60]{
\begin{tabular}{r|rrrrrrrrrrrrrrrrrrrrr}
        &a1     &a2     &a3    &a4     &a5     &a6     &a7     &a8     &a9    &a10 &a11 &a12 &a13 &a14 &a15 &a16 &a17 &a18 &a19 &a20 &a21
        \\
\hline
a1     &      &.1277 &.1165 &.1892 &.1338 &.1196 &.1301 &.1536 &.1049 &.1741 &.2653 &.2509 &.3712 &.1489 &.1315 &.2345 &.1866 &.1375 &.1939 &.0868 &.2262
        \\
a2     &.1277 &      &.1160 &.2370 &.1543 &.1391 &.1173 &.1347 &.1617 &.1643 &.2845 &.1522 &.2817 &.1422 &.1464 &.1169 &.1114 &.1723 &.1861 &.1834 &.1606
        \\
a3     &.1165 &.1160 &      &.1668 &.0769 &.1119 &.1474 &.1353 &.1471 &.2447 &.2214 &.2015 &.3191 &.0990 &.0897 &.1919 &.1715 &.1186 &.1720 &.0958 &.1771
        \\
a4     &.1892 &.2370 &.1668 &      &.1468 &.1637 &.2363 &.3173 &.1133 &.3779 &.3533 &.4261 &.5766 &.2478 &.1402 &.3280 &.2332 &.2330 &.2460 &.1057 &.3028
        \\
a5     &.1338 &.1543 &.0769 &.1468 &      &.1128 &.1026 &.1214 &.1395 &.2114 &.2070 &.2094 &.3265 &.1582 &.0906 &.1677 &.1749 &.0940 &.1252 &.0964 &.1563
        \\
a6     &.1196 &.1391 &.1119 &.1637 &.1128 &      &.1256 &.1274 &.1184 &.1778 &.2264 &.2161 &.3179 &.1294 &.0754 &.1651 &.1132 &.0938 &.1834 &.1421 &.1254
        \\
a7     &.1301 &.1173 &.1474 &.2363 &.1026 &.1256 &      &.1131 &.1669 &.1856 &.1928 &.2188 &.3023 &.1735 &.1425 &.1223 &.1489 &.1190 &.1161 &.1835 &.1296
        \\
a8     &.1536 &.1347 &.1353 &.3173 &.1214 &.1274 &.1131 &      &.2087 &.1327 &.1907 &.1179 &.2089 &.1105 &.1108 &.0972 &.1221 &.1289 &.1721 &.2228 &.1041
        \\
a9     &.1049 &.1617 &.1471 &.1133 &.1395 &.1184 &.1669 &.2087 &      &.2454 &.3359 &.3063 &.4550 &.1769 &.1224 &.2264 &.1415 &.1986 &.2144 &.1320 &.2119
        \\
a10    &.1741 &.1643 &.2447 &.3779 &.2114 &.1778 &.1856 &.1327 &.2454 &      &.2371 &.1281 &.1843 &.1964 &.2129 &.1809 &.1558 &.1694 &.3145 &.3417 &.2009
        \\
a11    &.2653 &.2845 &.2214 &.3533 &.2070 &.2264 &.1928 &.1907 &.3359 &.2371 &      &.2189 &.2366 &.1738 &.2451 &.1890 &.2468 &.1972 &.2894 &.3544 &.2449
        \\
a12    &.2509 &.1522 &.2015 &.4261 &.2094 &.2161 &.2188 &.1179 &.3063 &.1281 &.2189 &      &.1226 &.1593 &.2342 &.1295 &.2077 &.2282 &.2843 &.3288 &.1815
        \\
a13    &.3712 &.2817 &.3191 &.5766 &.3265 &.3179 &.3023 &.2089 &.4550 &.1843 &.2366 &.1226 &      &.1923 &.3397 &.2185 &.2995 &.3113 &.5016 &.5017 &.2912
        \\
a14    &.1489 &.1422 &.0990 &.2478 &.1582 &.1294 &.1735 &.1105 &.1769 &.1964 &.1738 &.1593 &.1923 &      &.1444 &.1752 &.1514 &.1907 &.2673 &.1665 &.1432
        \\
a15    &.1315 &.1464 &.0897 &.1402 &.0906 &.0754 &.1425 &.1108 &.1224 &.2129 &.2451 &.2342 &.3397 &.1444 &      &.1586 &.1224 &.1159 &.1655 &.1307 &.1629
        \\
a16    &.2345 &.1169 &.1919 &.3280 &.1677 &.1651 &.1223 &.0972 &.2264 &.1809 &.1890 &.1295 &.2185 &.1752 &.1586 &      &.1392 &.1809 &.1737 &.2762 &.1106
        \\
a17    &.1866 &.1114 &.1715 &.2332 &.1749 &.1132 &.1489 &.1221 &.1415 &.1558 &.2468 &.2077 &.2995 &.1514 &.1224 &.1392 &      &.1881 &.2208 &.1880 &.1372
        \\
a18    &.1375 &.1723 &.1186 &.2330 &.0940 &.0938 &.1190 &.1289 &.1986 &.1694 &.1972 &.2282 &.3113 &.1907 &.1159 &.1809 &.1881 &      &.1668 &.1922 &.1545
        \\
a19    &.1939 &.1861 &.1720 &.2460 &.1252 &.1834 &.1161 &.1721 &.2144 &.3145 &.2894 &.2843 &.5016 &.2673 &.1655 &.1737 &.2208 &.1668 &      &.1756 &.1438
        \\
a20    &.0868 &.1834 &.0958 &.1057 &.0964 &.1421 &.1835 &.2228 &.1320 &.3417 &.3544 &.3288 &.5017 &.1665 &.1307 &.2762 &.1880 &.1922 &.1756 &      &.2233
        \\
a21    &.2262 &.1606 &.1771 &.3028 &.1563 &.1254 &.1296 &.1041 &.2119 &.2009 &.2449 &.1815 &.2912 &.1432 &.1629 &.1106 &.1372 &.1545 &.1438 &.2233 &     
\end{tabular} 
}
}
\caption{\small Dissimilarity table for the \texttt{a}}  \label{diss.a.completas}
\end{table*}

\subsection{Dendrograms for letter \texttt{a}}\label{dendrosA}

\begin{figure*}
    \centering  
    {\includegraphics[scale=0.7]{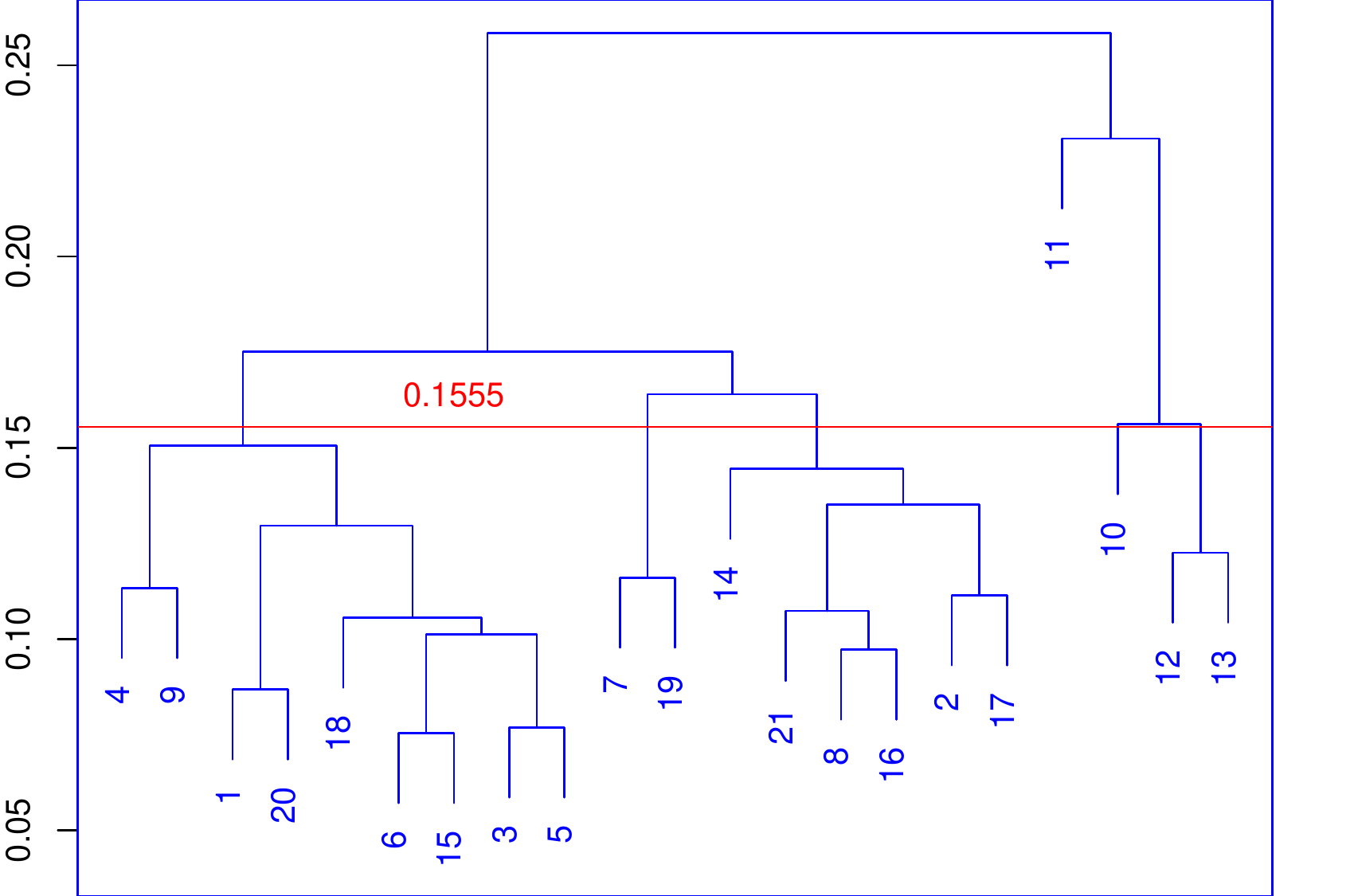}}  
    \caption{\small Dendrogram for letter \texttt{a} with the \emph{average linkage} method}\label{hclust_a_average-line}
    \end{figure*}

The dendrogram corresponding to the average linkage variant of the hierarchical clustering
is displayed in Figure \ref{hclust_a_average-line}.
  
  The order in which the different clusters are sorted horizontally does not have any special meaning.
  Here we follow the convention of drawing to the left the more compact cluster, that is, the 
  one formed at the lowest level.
      The corresponding sequence of cluster formation, sorted by merging level, is given in the following list.
      We follow the same convention of the analogous list above. 
      
\par      
\medskip
\begin{center}
{\ttfamily
        \setlength{\extrarowheight}{2pt}
      \begin{tabular}{c|c|c|}
       \cline{2-3} 
         & {\rmfamily merge}  & {\rmfamily distance} \\
      \whline
          \multicolumn{1}{|c|}{a}  &   ~6 --   15 &   0.0754      \\
          \multicolumn{1}{|c|}{b}  &   ~3 --   5~ &   0.0769      \\
          \multicolumn{1}{|c|}{c}  &   ~1 --   20 &   0.0868      \\
          \multicolumn{1}{|c|}{d}  &   ~8 --   16 &   0.0972      \\
          \multicolumn{1}{|c|}{e}  &    a --    b &   0.1012      \\
          \multicolumn{1}{|c|}{f}  &   18 --   e~ &   0.1056      \\
          \multicolumn{1}{|c|}{g}  &   21 --   d~ &   0.1074      \\
          \multicolumn{1}{|c|}{h}  &   ~2 --   17 &   0.1114      \\
          \multicolumn{1}{|c|}{i}  &   ~4 --   9~ &   0.1133      \\
          \multicolumn{1}{|c|}{j}  &   ~7 --   19 &   0.1161      \\
          \multicolumn{1}{|c|}{k}  &   12 --   13 &   0.1226      \\
          \multicolumn{1}{|c|}{l}  &   c  --    f &   0.1296      \\
          \multicolumn{1}{|c|}{m}  &   g  --    h &   0.1351      \\
          \multicolumn{1}{|c|}{n}  &   14 --   m~ &   0.1445      \\
          \multicolumn{1}{|c|}{o}  &    i --    l &   0.1506      \\
          \multicolumn{1}{|c|}{p}  &   10 --   k~ &   0.1562      \\
          \multicolumn{1}{|c|}{q}  &    j --    n &   0.1640      \\
          \multicolumn{1}{|c|}{r}  &    o --    q &   0.1752      \\
          \multicolumn{1}{|c|}{s}  &   11 --   p~ &   0.2309      \\
          \multicolumn{1}{|c|}{t}  &    r --    s &   0.2584      \\
      \hline                                                          
       \end{tabular}
  }
\end{center}

\par
\medskip
      Concerning the choice of the clustering in the hierarchy, which amounts to determine the 
      number of clusters, the usual heuristic, when no other information
      is available, is to  ``cut'' the dendrogram wherever the jumps in the distances of consecutive 
      mergings are higher than a threshold magnitude.                    
      For instance, the jump from 0.1752 to 0.2309 between lines \texttt{r} and  
      \texttt{s} would give us a partition in three clusters that looks quite clear.
            
      But we do have an additional information to determine the number and composition of clusters.
      This information is given by a set of dissimilarities measured from a test printing of 
      newly created types. We have used types belonging to the typographic collection of the 
      Bauer Neufville Type Foundry, kept at the Universitat de Barcelona. 
      These will play the role of ``control group'', since we know for sure
      that they come from the same matrix. If we try to divide them in clusters, all of them
      must therefore become members of the same and only cluster.
      Comparing these letters in the same way as we did with the microfilmed letters of the 
      Gospel of Matthew, we can see the magnitude of the 
      dissimilarities that we can expect and that are attributable only to the  errors of construction of the 
      metal types and to printing errors.
      
      The actual size of the types of the control group is of course different from the one of the 
      microfilmed letters, 
      so that the absolute values of the dissimilarities cannot be directly used to compare 
      the two groups.
      We have to define a common comparison base that will not be influenced
      by the different size. A natural and easy way to establish a common unit of measure is the following:
      
    \begin{enumerate}
    \item
      The control group is organised around a model letter, with the same procedure
      used with letters \texttt{i} in Section \ref{clustersI} (all of them in the same cluster,
      obviously),
      obtaining a set of distances to the model.
    \item  
      The quotient is computed between the largest and the smallest distance to the model,
      as a measure of dispersion of these distances, which is a value independent of the units
      of measure.
    \end{enumerate}  
    
      We have applied this procedure to the letters \texttt{m},
      \texttt{i}, \texttt{a}, \texttt{o}, with five specimens each, and the results are summarised in 
      Table \ref{tablapol}.
      
  \begin{table}
\centering  
    {\small  
        \setlength{\extrarowheight}{2pt}
      \begin{tabular}{r|c|c|c|c|}
      \cline{2-4}  
      \textrm{} & \textrm{distances to model} 
      & \textrm{mean} & \textrm{max/min} \\
      \whline 
       \multicolumn{1}{|c|}{\texttt{m}} 
       & 5.9731,\ 3.9409,\ 5.5055,\ 3.9967 & 4.8540 & 1.5156  \\
      \hline 
       \multicolumn{1}{|c|}{\texttt{i}} 
       & 3.2364,\ 5.9557,\ 2.8876,\ 3.9188 & 3.9996 & 2.0625   \\
      \hline
       \multicolumn{1}{|c|}{\texttt{a}} 
       & 7.4534,\ 4.8179,\ 4.6993,\ 3.7450 & 5.1789 & 1.9902   \\
      \hline
       \multicolumn{1}{|c|}{\texttt{o}} 
       & 1.7965,\ 1.8714,\ 2.0286,\ 1.7940 & 1.8726 & 1.1307  \\
      \hline
       \end{tabular}
    }  
  \caption{\small Distances obtained with brand new metal types, and five specimens per glyph}\label{tablapol}
  \end{table}     
      
      The quotient between the largest and the smallest distance to the model
      gives us an estimate of the variability that we can expect
      from a unique matrix. We see that the quotients range between 1.1307 and 2.0625.
      
      We will use this estimate in the following way for our set of \texttt{a}: We take 
      the two letters which are the closest and declare them, obviously, as belonging to the same cluster.
      This minimum distance is 0.0754, and it is found between letters \texttt{a6} and \texttt{a15}. 
      According to the data obtained from the control group, the letters at a distance up to 
      2.0625 times this minimum distance 
      (i.e. distances up to 0.1555) may perfectly belong to the same cluster.
      
      Therefore, we can draw a cutting line in the dendrogram 
      at level 0.1555 (the red line in Figure \ref{hclust_a_average-line}), 
      and declare all clusters grouped below this line
      as indivisible.

      Thus, we obtain six clusters. We can now construct the stars corresponding to these clusters, 
      finding their model letters and the distances from the remaining letters to its model 
      (see Table \ref{clust_a_prelim}).
      
\begin{table*}
\centering  
        \setlength{\extrarowheight}{2pt}
      \begin{tabular}{c|r|c|c|c|c|}
      \cline{2-6}  
       & \textrm{cluster}  & \textrm{model} & \textrm{mean of distances} & \textrm{minimum, maximum}&\textrm{max/min} \\
      \whline 
        \multicolumn{1}{|c|}{1} &\{4,9,1,20,18,6,15,3,5\} & 5 & 0.1114 &  0.0769, 0.1468 & 1.9090\\
      \hline
        \multicolumn{1}{|c|}{2} &\{7,19\} & 7, 19 & 0.1161 & &  \\
      \hline
        \multicolumn{1}{|c|}{3} &\{14,21,8,16,2,17\} & 8 & 0.1137 &  0.0972, 0.1347 &1.3858\\
      \hline
        \multicolumn{1}{|c|}{4} &\{11\} & 11 &  - &  &\\
      \hline
        \multicolumn{1}{|c|}{5} &\{10\} & 10 & - &  &\\
      \hline
        \multicolumn{1}{|c|}{6} &\{12,13\} & 12, 13 & 0.1226 & & \\
      \hline                                                          
       \end{tabular}
\caption{\small Preliminary clustering for letter \texttt{a}}\label{clust_a_prelim}      
\end{table*}

    Our conclusion up to this moment is that the maximum number of clusters that we obtain from the  
    twenty-one letters \texttt{a} is six, because splitting any one of them would be analogous to declare
    that the metal types of the studied control group came from more than one matrix, which is false. 
    
    \subsection{Validation of clusters}\label{validclust}
    
      We are now interested in checking if we have actually gone too far 
      in the number of clusters found, and some of them should better be merged
      into larger ones, following our conservative criterion
      to declare two letters as coming from the same matrix
      if there is no strong evidence against this hypothesis.
      Take also  into account that the dendrogram is the result of an approximate method,
      built upon a somewhat arbitrary choice of the criterion with which the hierarchy is 
      constructed; it is therefore absolutely necessary to validate the present clusters 
      with some sieve reflecting the conservative spirit. 
    
      To this end, we use a statistical test for the detection of outliers. In general 
      an \emph{outlier} in a set of numerical data is a datum markedly different from the rest,
      that seems to result from some sort of measuring error or typo, or in summary 
      that \emph{should not be there}, because it does not really belong to that data set. 
      
      As an example, let us take cluster number 3. 
      The model is letter \texttt{a8} and so we can draw a star with center in  
      \texttt{a8} and the corresponding distance to the other letters. These distances are,  
      sorted in increasing order: 
      
\centerline{0.0972  0.1041  0.1105  0.1221 0.1347}
      
      Take now any other letter, not belonging to cluster 3, and observe its distance to the model \texttt{a8}. 
      For instance,  \texttt{a1} happens to be at a distance 0.1536 from \texttt{a8}. 
      Is this distance too large to include also \texttt{a1} in cluster 3? 
      Or, on the contrary, it is not much different from the others, and the larger distance is only 
      the product of chance?
      
      The conservative hypothesis, and therefore the one that it is considered true a priori, as in any statistical 
      test, is the second possibility above: the distance between \texttt{a1} and \texttt{a8} is not 
      exceedingly large 
      and in consequence \texttt{a1} should also belong to cluster 3. This \emph{null hypothesis} will be abandoned  
      only if there is enough evidence against it. The evidence is measured in terms of the $p$-value,  
      the probability of rejecting the null hypothesis when it was in fact true. If the $p$-value is less than a certain threshold 
      fixed beforehand (typically 10\%, 5\% or 1\%), then the null hypothesis is rejected; otherwise,
      it is kept, conservatively, as the good one (albeit with a probability to err with this conclusion
      that can be very high).
      
      We are going to apply the so-called \emph{Dixon's test}, the usual one to test for the  
      presence of one outlier in a data set. 
      Let $x_1,\dots,x_n$ be the data set, sorted in increasing order, 
      including as $x_n$ the suspect datum. Dixon's test computes the statistic 
\begin{equation*}
      Q:=\frac{x_n-x_{n-1}}{x_n-x_1}
      \ .
      \end{equation*}      
      Intuitively, a large value for this quotient indicates 
      that there is a large difference between the last value and the one before last
      (relatively to the dispersion of the whole data set), inducing us to believe      
      that the last value is an outlier and to reject the working hypothesis. If, on the contrary, 
      the quotient is small, then the difference between $x_{n-1}$ and $x_n$ is small
      and there is no reason to declare $x_n$ as outlier.
      
      Continuing the example above, if we want to see if letter \texttt{a1} belongs to cluster 3,    
      we compute the value of the Dixon's statistic:
\begin{equation*}
      Q=\frac{0.1536-0.1347}{0.1536-0.0972}=0.3351
      \ .
\end{equation*}      
      
      In order to apply reliably Dixon's test to determine if the value 
      0.3351 is too large or not, the data must satisfy two assumptions that in our 
      case are fulfilled: 
\begin{enumerate} [a)]
\item 
      The data follow a Gaussian law, and
\item  
     they are statistically independent.
 \end{enumerate}      
  
  Gaussianity is justified by the fact that each datum
  is the sum of a large quantity of very small values (the distances from points of one contour to 
  segments of the other, divided by its number, see Section \ref{lab}). Therefore, 
  independently of the underlying 
  probability law of those small values, the sum will very approximately follow a Gaussian law,  
  thanks to the Central Limit Theorem.
  
  The independence is obvious because the result of measuring the distance form any letter  
  to the ``model'' letter does 
  not influence the result of the distance between a third letter and the model. (Notice that
  if we include the distance among the letters which are not models, the independence  
  would be lost.)
         
  Under these conditions, the probability distribution of the random variable $Q$,     
  assuming that there are no outliers, is known. This allows to compute the probability
  that $Q$ attain a value greater than or equal to the observed one   
  if $x_n$ is not an outlier. In the case of our example, the probability of the event
  $\{Q\ge0.3351\}$ is approximately 0.27. This is the probability of mistakenly declare
  that $x_n$ is an outlier. Of course, it is too high to take the risk.
      
  Letter \texttt{a1} may therefore be included in cluster 3. But we have said that cluster 1
  will not be split. So, let us check what happens with the remaining letters in cluster 1,
  namely \texttt{a4}, \texttt{a9} and \texttt{a20}, which are at distances 
  0.3173, 0.2087, 0.2228 from the model \texttt{a8} of cluster 3 (see Table \ref{diss.a.completas}). 
  Using Dixon's test one finds that all three of them fall below the 5\% threshold of probability of
  a mistake if we declare them outside cluster 3 ($<0.0001$, 0.0156, and 0.0095, respectively). 
  So this is what we do and we keep definitively separated clusters 1 and 3.
      
  Repeating the same idea with the remaining clusters, we obtain that:
\begin{itemize}
\item 
  Cluster 5 integrates in cluster 6 (in fact, they were close to be already merged when 
  we drew the horizontal line in the dendrogram).
\item  
  Cluster 2 can easily be integrated in cluster 1. In this case, both distances to the model
   \texttt{a5} are even smaller than those of some other letters already included in cluster 1. 
  Cluster 2 could also be integrated in cluster 3, but not so clearly, since 
  the $p$-value for letter \texttt{a19} is 0.0860, so that with the milder threshold of  
  10\% as error probability it will not be integrated.
\item
  Cluster 4 (letter \texttt{a11}) does not integrate in the new cluster union of 5 and 6 
  (letters \texttt{a10}, \texttt{a12}, \texttt{a13}),
  with $p$-value $0.0485$.
  Similarly, it cannot be integrated in the cluster union of 1 and 2 (with $p$-value $0.0193$), nor
  in cluster 3 ($p$-value $0.0332$). 
  Thus letter \texttt{a11} remains alone definitively.
\item  
  It can be checked that there are no other possibilities for merging the new clusters among them.
\end{itemize}  

We arrive then to the final classification in four clusters given in Table \ref{clusterdef}.

\begin{table*}
\centering  
    \setlength{\extrarowheight}{2pt}
  \begin{tabular}{c|c|c|c|c|c|}
  \cline{2-6}  
    & \textrm{cluster}  & \textrm{model} & \textrm{mean of distances} & \textrm{minimum, maximum}&\textrm{max/min} \\
  \whline 
    \multicolumn{1}{|c|}{A} &\{4,9,1,20,18,6,15,3,5,7,19\} & 5 & 0.1114 &  0.0769, 0.1468 & 1.9090\\
  \hline
    \multicolumn{1}{|c|}{B} &\{14,21,8,16,2,17\} & 8 & 0.1137 &  0.0972, 0.1347 &1.3858\\
  \hline
    \multicolumn{1}{|c|}{C} &\{11\} & 11 &  - &  &\\
  \hline
    \multicolumn{1}{|c|}{D} &\{10,12,13\} & 12 & 0.1254 & 0.1225, 0.1281 & 1.0457\\
  \hline
   \end{tabular}
\caption{\small Final classification in clusters for of letters \texttt{a}}\label{clusterdef}  
\end{table*}

  We remark once again that the procedure we have used is conservative: 
  we are pretty sure that there is more than one cluster (four at least) in the 21 letters \texttt{a}
  studied, but it is perfectly possible that there are actually more than 4.

  We also remark that sometimes an outlier can be masked by the presence of another questionable datum.
  There exist statistical tests to expose the existence of two outliers at once (e.g. Grubbs' test,
  see \cite{Barnett}); however, 
  following our conservative spirit, we have preferred not to try to identify such situations and, if
  it happens, to accept both data as genuine.
  
  At first sight, the decision of declaring as separate two clusters when just one of the letters of the first 
  cluster is rejected as admissible by the second one may seem more risky. But one should take into account 
  that
  it is still more risky to split a cluster which has been first declared indivisible. Suppose, on the other hand,
  that we do not want to split the first cluster but we want to know if it should be completely integrated in the 
  second. Recall that this must be the null hypothesis of the test. Now, the probability of error
  in declaring mistakenly out of a cluster more than one letter is still smaller than the probability
  of declaring out of the cluster each one of them. 
  Therefore, if one of the letters does not exceed the fixed 5\% of probability of 
  error, a set of more than one letter will not exceed that value either.
  
  The following are the computed probabilities that have allowed to conclude 
  the final classification of Table \ref{clusterdef} from the preliminary 
  classification of Table \ref{clust_a_prelim},
  through Dixon's test. 
  For each 
  possible destination cluster, we list the candidate letters, with the
  cluster to which they belong.
  Only the rejections ($p$-values less than 5\%) are shown. 
  $1\unio 2$ and $5\unio 6$ mean the union of these clusters, whose model letters
  turn out to be \texttt{a5} and \texttt{a12}, respectively.
  
\par  
\medskip
\begin{center}
\scalebox{.85}[.85]{
    \setlength{\extrarowheight}{2pt}
  \begin{tabular}{|c|c|c|r|}
    \textrm{letter}  & \textrm{from cluster} & \textrm{to cluster} & \textrm{$p$-value\ } \\
  \whline 
     10   &      5  & 1 &  0.0294    \\
     11   &      4  & 1 &  0.0368    \\
     12   &      6  & 1 &  0.0326    \\
     13   &      6  & 1 & $<0.0001$  \\
  \hline   
     4    &      1  & 3 &$<0.0001$   \\
     9    &      1  & 3 &  0.0156    \\
    11    &      4  & 3 &  0.0332    \\
    13    &      6  & 3 &  0.0155    \\
    20    &      1  & 3 &  0.0095    \\   
  \hline
    10   &       5  & $1\unio 2$ &  0.0146    \\  
    11   &       4  & $1\unio 2$ &  0.0193    \\  
    12   &       6  & $1\unio 2$ &  0.0167    \\  
    13   &       6  & $1\unio 2$ &  $<0.0001$ \\      
  \hline
     1   &      1  & $5\unio 6$  & 0.0360     \\ 
     6   &      1  & $5\unio 6$  & 0.0500     \\ 
     7   &      2  & $5\unio 6$  & 0.0485     \\ 
     9   &      1  & $5\unio 6$  & 0.0250     \\ 
    11   &      4  & $5\unio 6$  & 0.0485     \\
    15   &      1  & $5\unio 6$  & 0.0416     \\ 
    18   &      1  & $5\unio 6$  & 0.0441     \\ 
    19   &      2  & $5\unio 6$  & 0.0284     \\ 
    20   &      1  & $5\unio 6$  & 0.0223     \\ 
   \hline                             
   \end{tabular}
   }
\end{center}

\par  
\medskip
  To close the section, we return for a moment to the analysis of the ten letters 
  \texttt{i}, in connection with the validation of clusters. 
  
  In Section \ref{clustersI}, we obtained the optimal (exact) clusterings,    
  conditioned to the existence of exactly two or three clusters. We can also apply
  Dixon's test for outliers to those clusterings, and the result confirms the partitions
  seen. It is also natural to use the test to check if a partition in 4 clusters
  could be reasonable. The result is negative:
  
  The best clustering with four clusters (exact, with the criterion introduced in
  Section \ref{clustersI}) is 
\begin{equation*}
  \alpha=\{1\}\ ,\quad
  \beta=\{3\}\ ,\quad
  \gamma=\{5\}\ ,\quad
  \delta=\{2,4,6,7,8,9,10\}\ .
\end{equation*}  
  But Dixon's test does not support \texttt{i5} out of cluster $\delta$. It does
  for \texttt{i1} and \texttt{i3}, with $p$-values $<0.0001$ and 0.0117, respectively.    
 
The second best clustering with four clusters corresponds to     
\begin{equation*}
  \alpha=\{1\}\ ,\quad
  \beta=\{3\}\ ,\quad
  \gamma=\{7\}\ ,\quad
  \delta=\{2,4,5,6,8,9,10\}\ .
\end{equation*}  
which looks more consistent with our previous partition in three clusters. 
Again \texttt{i1} and \texttt{i3} cannot be integrated in $\delta$, with  
$p$-values 0.0032 and 0.0075 respectively, 
and, although \texttt{i7} could be, we should integrate first \texttt{i3}, from the dendrogram, 
given that we cannot perform an outliers test against a cluster with one only element.
     
As a curiosity, it turns out that the model letters (center of the star) for clusters $\delta$ 
in the two cases above are 
different. They are \texttt{i8} in the first case and \texttt{i2} in the second. In fact, they 
are very close letters and 
these changes are not surprising
when changing slightly the composition of a cluster.

In conclusion, this study forcing four clusters does not bring us more than what was already 
shown: We can postulate with confidence the existence of three clusters, and 
concerning its composition, the safest bet is our first option
\begin{equation*}
  \alpha=\{1\}\ ,\quad
  \beta=\{3,7\}\ ,\quad
  \delta=\{2,4,5,6,8,9,10\}\ ,
\end{equation*}  
with a certain risk, coarsely bounded by a probability of 0.32 of error in separating letter 7
from cluster $\delta$.  The clustering with two clusters, merging $\beta$ and $\delta$, looks to us
extremely conservative.

\section{Graphical representation of clusterings} \label{MDS}

  The \emph{Multidimensional Scaling} (MDS) is a technique that allows to represent
  graphically in dimensions 2 or 3 the set of objects that are being classified,
  providing a visual impression of the closeness among different objects or clusters of objects.
  The distances in the display are not the true dissimilarities;
  they are the best possible approximation (according to some loss function) so that the
  data can be embedded in the 2- or 3-dimensional space.
  
  The specific coordinates of the points have no intrinsic meaning;
  only the relative position among objects are to be read.
  In particular, any rotation, symmetry or translation of the picture
  gives rise to another picture with exactly the same meaning.
  Yet it can be assured that the mapping is monotone: if an object possesses 
  dissimilarities $d_1$ and $d_2$ with other
  two objects, and $d_1<d_2$, then the corresponding distances in the picture
  $\hat d_1$ and $\hat d_2$ also satisfy $\hat d_1<\hat d_2$. 
  
\begin{figure*}
\centering  
{\includegraphics[scale=0.60]{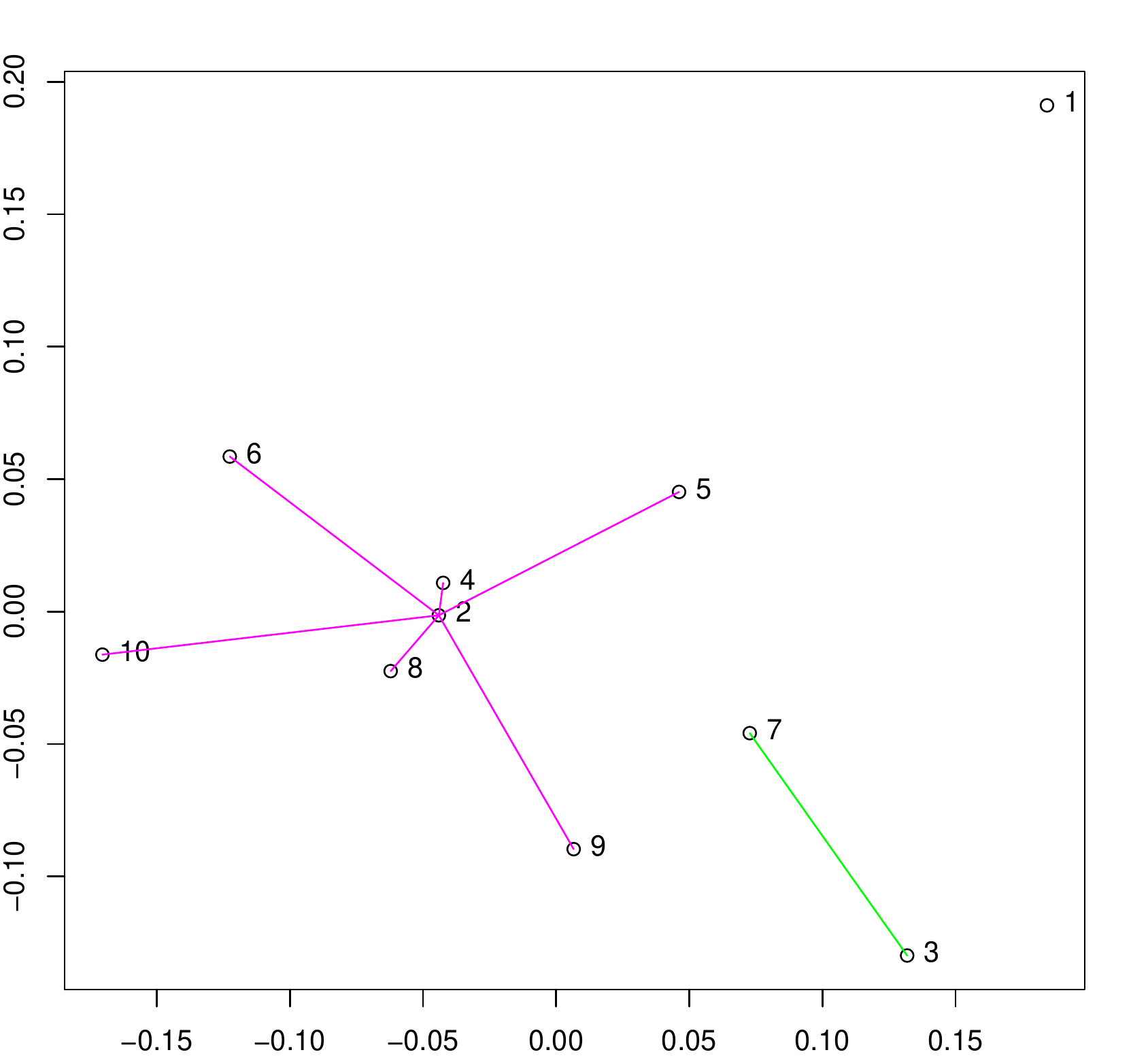}}  
\caption{\small MDS in dimension 2 for letter \texttt{i}}\label{MDS_i_isoMDS_2d}
\end{figure*}

\begin{figure*}
\centering  
{\includegraphics[scale=0.60]{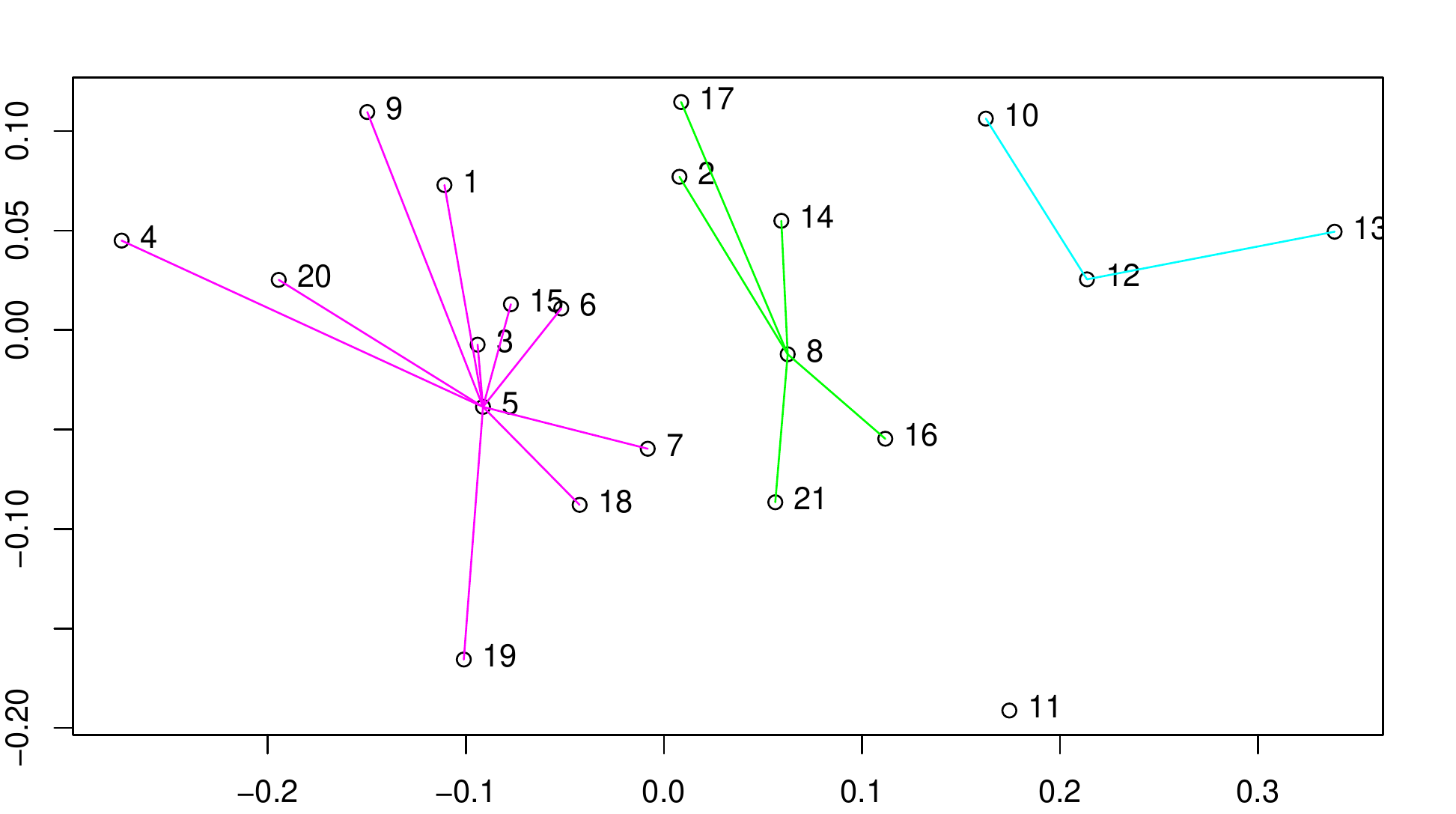}}  
\caption{\small MDS en in dimension 2 for letter \texttt{a}}\label{MDS_a_isoMDS_2d}
\end{figure*}

  The MDS in dimension 2 for the 10 letters \texttt{i} and the 21 letters \texttt{a} are shown
  in Figures \ref{MDS_i_isoMDS_2d} and \ref{MDS_a_isoMDS_2d}. 
  The stars of the optimal clustering found have also been displayed, and we see that the clusters
  appear quite clearly in the pictures.

  The extent to which a possible graphical representation of 
  a set of objects differs from the real dissimilarities is
  called \emph{stress} and can be defined in several reasonable ways.
  The representation used here corresponds to the variant known as 
  \emph{Kruskal Non-metric Multidimensional
  Scaling}. The stress $S$ is defined as 
\begin{equation*}
  S=\sqrt{\frac{\sum (f(d_{ij})-\hat d_{ij})^2}{\sum \hat d_{ij}^2}}
\end{equation*}  
  where $d$ are the given dissimilarities among the objects, $\hat d$ are the distances 
  in the graphic, and $f$ is a monotonically increasing transformation that gives an extra flexibility
  to the adjustment.
  The position of the points are obtained by seeking the distances $\hat d$ and the transformation $f$
  that minimize the stress.   
  The monotonicity of $f$ ensures that if one dissimilarity is smaller than another, 
  then the corresponding distances in the picture preserve that order. 
  
  More details on Kruskal Non-metric Multidimensional Scaling can be seen, for instance,
  in Chapter 3 of \cite{Cox}, or in Chapter 11 of \cite{Ripley-Venables}. 

   
\section{Conclusion and open questions}\label{conclusion}     

  From the historical research viewpoint, finding out which tools were used or developed
  by Gutenberg and Sch÷ffer to complete their monumental work is one of the most
  interesting items. 
  The fact that metal types built from multiple matrices
  were used concurrently, as we prove here, is a new aspect that indicates that matrix 
  construction was quite evolved. The use of punches and counterpunches 
  to sculpt the inner contours is agreed today;
  the tools used for the outer
  contours are not so clear. The letters \texttt{a}, and other letters with counters,
  offer the additional possibility to shed light in this respect.

  We have carried out a prospective study with the inner lower contour of 
  letter \texttt{a} (in the \emph{gothic textura} typeface it has indeed two inner contours), 
  and the results seem to indicate that different outer 
  contours combine with different inner contours. This may imply that 
  punches were also used to delineate the external shape of the matrix.
  However, we feel that we should first gather many more experimental 
  data, implementing also an improved measuring protocol, and we have therefore decided
  not to include those results here.
   
  We have also started to measure letters coming from the subsequent pages 
  of the same bible book, the Gospel of Matthew. We have computed their distances
  with the twenty-one letters of the first page, just to see if they can be 
  integrated into the existing clusters or not. What we observe is that most 
  of them integrate well in the clusters, but that a few new ``models'' appear. 
  This was absolutely in line with what we expected after obtaining the results 
  concerning the first page. 

  Notice that, using letters coming from different pages, and that may therefore 
  be printed from the same metal type, three levels of similarity should be observed:  
  Letters very similar among them, appearing for sure in different pages, printed with
  the same metal type;
  similar letters, but not that much, printed with different types but with a common
  matricial origin; and letters with low similarity, printed from types coming 
  from different matrices. We do not have yet enough data to confirm this, 
  but we believe it is feasible to carry out this programme in the future. 

\par
\bigskip
  \textbf{Acknowledgements.} We thank Prof. Enric Tormo from the Universitat de Barcelona for his guidance in the historical
  aspects. 
  
  None of the authors is supported by any grant of the Spanish or Catalan governments.


\end{document}